\documentclass{article} 
\usepackage[preprint]{colm2025_conference}

\usepackage{microtype}
\usepackage{hyperref}
\usepackage{url}
\usepackage{graphicx}
\usepackage{amsmath}
\usepackage{multirow}
\usepackage{booktabs}
\usepackage{times}
\usepackage{latexsym}
\usepackage{colortbl, xcolor}
\usepackage{paralist}

\usepackage{lineno}

\definecolor{darkblue}{rgb}{0, 0, 0.5}
\hypersetup{colorlinks=true, citecolor=darkblue, linkcolor=darkblue, urlcolor=darkblue}

\title{Feeding LLM Annotations to BERT Classifiers at Your Own Risk}
\author{Yucheng Lu\thanks{New York University. Email: yl6586@nyu.edu}
\And Kazimier Smith\thanks{New York University. Email: kas1112@stern.nyu.edu}}

\begin{document}

\ifcolmsubmission
\linenumbers
\fi

\maketitle

\begin{abstract}
Using LLM-generated labels to fine-tune smaller encoder-only models for text classification has gained popularity in various settings. While this approach may be justified in simple and low-stakes applications, we conduct empirical analysis to demonstrate how the perennial curse of training on synthetic data manifests itself in this specific setup. Compared to models trained on gold labels, we observe not only the expected performance degradation in accuracy and F1 score, but also increased instability across training runs and premature performance plateaus. These findings cast doubts on the reliability of such approaches in real-world applications. We contextualize the observed phenomena through the lens of error propagation and offer several practical mitigation strategies, including entropy-based filtering and ensemble techniques. Although these heuristics offer partial relief, they do not fully resolve the inherent risks of propagating non-random errors from LLM annotations to smaller classifiers, underscoring the need for caution when applying this workflow in high-stakes text classification tasks.
\end{abstract}

\section{Introduction}
\label{sec:intro}
Text classification remains a crucial application of LLMs. In settings where unlabeled data is abundant but gold labels and computational resources are scarce, recent work (e.g., \citet{golde2023fabricator, pangakis-wolken-2024-knowledge, 10825193}) suggested fine-tuning smaller encoder-only language models, such as BERT \citep{devlin-etal-2019-bert} using LLM-generated labels as training samples. This strategy promises to strike a balance between performance and cost, and has become increasingly popular across commercial, academic, and policy applications, some of which carry potentially high societal impact. Examples range from healthcare \citep{Kumichev_2024, smolyak2024large} to legal analysis \citep{freitas-2024-text, colombo2024saullm7bpioneeringlargelanguage}, and to policy decision making \citep{dell2024deeplearningeconomists, halterman2025codebookllmsevaluatingllms}. 

However, the reliability of such approaches remains under-explored. Previous work often treats LLM-generated labels as adequate approximations of human annotations, focusing narrowly on performance parity (\citet{wang-etal-2021-want-reduce, csanády2024llambertlargescalelowcostdata, pangakis2024keepinghumansloophumancentered}). This overlooks risks inherent to synthetic data training, such as error propagation and model collapse—issues well-documented in broader machine learning literature \citep{bauer2024comprehensiveexplorationsyntheticdata, liu2024best, shumailov2024curserecursiontraininggenerated}. 
These gaps are particularly consequential in applied settings like computational social science, where researchers increasingly leverage LLM annotations for large text corpora despite lacking validation mechanisms \citep{https://doi.org/10.1111/ajps.12875}. While prior work has studied synthetic text-label pairs \citep{kuo2024llmgenerateddataequalrethinking, li-etal-2023-synthetic}, our focus on label generation alone addresses a more common real-world constraint: abundant unlabeled text data paired with expensive annotation processes.

We address this gap through experiments on four benchmark datasets of varying complexity, demonstrating that the trade-offs of training with LLM-generated labels extend beyond modest accuracy/F1 degradation. In summary, our main contributions are:

\begin{enumerate}
    \item \textbf{Empirical Analysis of Synthetic Label Training:}  We reveal how synthetic labels erode prediction robustness and leads to early performance plateau —  dimensions often ignored in prior analyses. These phenomena persist across datasets, contradicting assumptions of "more data always helps."
    \item \textbf{Evaluation of Mitigation 
    Strategies:} We test entropy-based filtering (removing low-confidence LLM labels) and consistency ensembles (aggregating multiple LLM annotations), showing they recover only 60–75\% of the gold-label performance gap. More critically, neither strategy stabilizes training variance or mitigates early plateaus, underscoring fundamental limitations of post hoc corrections.
\end{enumerate}
Our findings challenge the premise that synthetic labels are a "safe" substitute for human annotations, even in ostensibly simple classification tasks. The paper proceeds as follows: Section \ref{sec:baseline setup} details baseline experimental protocols, while Section \ref{sec:baseline results} analyzes performance degradation, instability, and performance plateau alongside theoretical interpretation. Sections \ref{sec:mitigation setup} and \ref{sec:mitigation results} evaluate mitigation strategies, concluding with implications for practitioners relying on LLM-generated training data.

\section{Baseline Experiments}
\label{sec:baseline setup}
\subsection{Methods}
We compare classifiers fine-tuned on LLM-generated labels vs. gold labels across four datasets chosen for task diversity and difficulty: \begin{itemize}
    \item \textbf{IMDB}: balanced binary sentiment analysis
    \item \textbf{ECommerce}: slightly imbalanced multi-class product categorization
    \item \textbf{Manifestos}: nuanced political stance detection, imbalanced data, smaller training size
    \item \textbf{Toxic}: hate speech vs. offensive language detection on twitter texts, highly imbalanced
\end{itemize} 
Details about these datasets are in Appendix \ref{sec:datasets}. Following prior work, we use \texttt{roberta-base} with standard classification heads as our encoder-only classifiers. For annotation, we use \texttt{Qwen2.5-Instruct} (3b, 7b), as representatives LLMs in their respective weight classes \citep{qwen2.5}. Using three-shot prompts, we generate synthetic labels for training texts while withholding gold labels. Fine-tuning details are Appendix \ref{sec:slm details}. Few shot classification details are in Appendix \ref{sec:llm details}. 

\subsection{Evaluation Metrics}
\paragraph{Accuracy and Macro-F1.}We evaluate the overall performance by looking at accuracy and macro-F1. Since we are especially interested in the stability of our classification models, we perform each experiment five times and compute the variance of accuracy and macro-F1 as well. 
\paragraph{Stability at the Individual Level.} In addition to variation in overall performances, another important indicator to consider in high-stakes situations is prediction stability at the individual level. We measure this using Krippendorff’s Alpha $\alpha_K$ which quantifies inter-rater agreement across training run and the proportion of unchanged predictions $ p_{uc}$ across five trials, providing an intuitive measure of model decisiveness.

\section{Baseline Results}
\label{sec:baseline results}
\paragraph{Non-Random Performance Degradation} Unsurprising, models trained on synthetic labels consistently underperform those trained on gold labels across all datasets, with the performance gap widening as task complexity increases. On IMDB, a simple benchmark first introduced in 2011, the performance difference is negligible. However, for multi-class classification on Ecommerce , models trained on labels from the 3B parameter LLM suffer a dramatic 30-point accuracy drop (66.05\% versus 96.26\% with gold labels). Notably, scaling up to a 7B parameter model fails to bridge this gap, achieving only 92.74\% accuracy.
The discrepancy between accuracy and F1 scores on the Manifestos and Toxic datasets reveals a more nuanced issue: LLM-generated labels lead to systematic failures in modeling tail distributions. Through manual error analysis, we found that both the LLM annotator and subsequently trained RoBERTa classifier consistently underperform on minority classes. This phenomenon can be interpreted as a mild form of model collapse during synthetic data training, as described by \citep{shumailov2024collapse}, where the model fails to adequately learn tail distributions.

\paragraph{Performance Plateau} As Figures 1 and 2 shows, models trained on synthetic labels exhibit premature performance plateaus compared to those trained on gold labels, showing diminishing returns as training data increases. The observed plateaus can be attributed to the propagation of systematic errors present in LLM annotations, as documented by \citep{chen2022pathologiespretrainedlanguagemodels} in few-shot learning contexts, as well as by \citep{li-etal-2023-synthetic}'s findings regarding LLMs' difficulties with subjective classification tasks. 

\paragraph{Prediction Instability Does Not Decrease with LLM Size} Perhaps the most concerning finding is that models trained on synthetic labels exhibit significant prediction instability, and this instability persists even when using larger LLMs for label generation. On the Manifestos dataset, we observe a dramatic drop in Krippendorff's alpha ($\alpha_K$) from 84.30 with gold labels to 52.72 with 3B labels, and this further deteriorates to 43.75 with 7B-generated labels. The proportion of unchanged predictions ($p_{uc}$) tells a similar story, dropping from 82.04\% to 50\% and 30.45\% respectively. This pattern holds across other datasets, though to varying degrees. Even on the simpler IMDB task, where accuracy remains competitive, we still see a consistent decline in prediction stability metrics. The Toxic dataset particularly highlights this issue, where using 7B-generated labels leads to high variance in predictions ($p_{uc}$ = 77.49\%) despite relatively strong accuracy scores. These results suggest that models trained on synthetic labels not only underperform but also make less consistent predictions across different training runs.

\begin{figure}[t]
    \centering
    \includegraphics[width=0.7\linewidth]{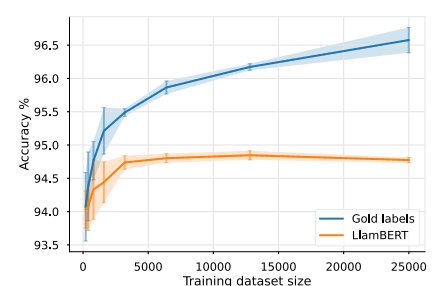}
    \caption{Performance of RoBERTa trained on "Gold labels" vs on synthetic labels ("LlamBERT"). Plot from \citet{csanády2024llambertlargescalelowcostdata}, as we limit training to 5000 data points to reduce environmental impact}
    \label{fig:enter-label}
\end{figure}

\begin{table*}[!ht]
\centering
\footnotesize
\setlength{\tabcolsep}{4pt}
\renewcommand\arraystretch{1.1}
\begin{tabular}{l l | ccc | ccc | ccc | ccc}
\hline
\multirow{3}{*}{Model} & \multirow{3}{*}{Label} & \multicolumn{3}{c|}{IMDB} & \multicolumn{3}{c|}{Ecommerce} & \multicolumn{3}{c|}{Manifestos} & \multicolumn{3}{c}{Toxic} \\
& & $\mu_{acc}$ & $\mu_{f1}$ & $\alpha_{K}$ & $\mu_{acc}$ & $\mu_{f1}$ & $\alpha_{K}$ & $\mu_{acc}$ & $\mu_{f1}$ & $\alpha_{K}$ & $\mu_{acc}$ & $\mu_{f1}$ & $\alpha_{K}$ \\
& & $\sigma_{acc}$ & $\sigma_{f1}$ & $p_{uc}$ & $\sigma_{acc}$ & $\sigma_{f1}$ & $p_{uc}$ & $\sigma_{acc}$ & $\sigma_{f1}$ & $p_{uc}$ & $\sigma_{acc}$ & $\sigma_{f1}$ & $p_{uc}$ \\
\hline
\multirow{6}{*}{\rotatebox{90}{RoBERTa-Base}}
& \multirow{2}{*}{Gold} & 93.84 & 93.82 & 90.6 & 96.26 & 96.23 & 96.69 & 83.56 & 79.38 & 84.30 & 91.13 & 75.24 & 84.08 \\
& & 0.28 & 0.29 & \cellcolor{red!60}90 & 0.25 & 0.22 & \cellcolor{red!60}95.24 & 0.69 & 0.54 & \cellcolor{red!60}82.04 & 0.23 & 1.01 & \cellcolor{red!60}88.50 \\
\cline{2-14}
& \multirow{2}{*}{3B} & 93.33 & 93.31 & 89.99 & 66.05 & 66.62 & 75.04 & 66.02 & 41.91 & 52.72 & 86.86 & 65.18 & 57.29 \\
& & 0.16 & 0.16 & \cellcolor{red!60}89.68 & 3.14 & 3.69 & \cellcolor{red!60}70.14 & 0.00 & 3.68 & \cellcolor{red!60}50 & 1.74 & 1.93 & \cellcolor{red!60}47.39 \\
\cline{2-14}
& \multirow{2}{*}{7B} & 92.95 & 92.94 & 87.28 & 92.74 & 92.88 & 79.18 & 71.51 & 60.62 & 43.75 & 83.89 & 56.53 & 68.49 \\
& & 0.20 & 0.20 & \cellcolor{red!60}86.56 & 0.81 & 0.76 & \cellcolor{red!60}69.35 & 1.30 & 7.96 & \cellcolor{red!60}30.45 & 5.31 & 5.04 & \cellcolor{red!60}77.49 \\
\hline
\end{tabular}
\caption{Experimental results across different models, label types, and datasets. For each dataset, we report average accuracy $\mu_{acc}$, average macro F1-score $\mu_{f1}$, standard deviation of accuracy $\sigma_{acc}$, and standard deviation of macro F1 $\sigma_{f1}$. In addition, we compute Krippendorff's alpha $\alpha_{K}$ and the proportion of predictions that remain unchanged across experimental runs $p_{uc}$. All numbers are scaled up by 100 for ease of presentation.}
\label{tab:baseline_results}
\end{table*}

\subsection{Theoretical Interpretation}
\paragraph{Framework} Denote the true data generating process of text and label pair as the joint distribution $P(Y, X)$, where $Y$ is label/class, and $X$ is input text. The supervised text classifier is trained to estimate the conditional distribution $P(Y|X)$ from i.i.d. sample $D_P=\{(y_i,x_i)_{i=1}^N\}$ by minimizing cross-entropy loss:
\begin{align*}
    \mathcal{L}_{\text{CE}}(\theta, D_P) = -\frac{1}{N} \sum_{i=1}^N \log \hat{P}(y_i|\mathbf{x}_i; \theta)
\end{align*}
However, since we are using labels generated from LLM, the data we see is actually drawn from \footnote{Incidentally, from this formulation, one can see that when a large pool of unlabeled text is available, using synthetic labels is theoretically superior than using synthetic text and label pairs, as it avoids additional LLM approximation error on the marginal distribution of input text $P(X)$. } 
\begin{align*}
  D_{S}=\{(y_i,x_i)_{i=1}^N\} & \sim P(X)P_S{(Y|X)}  \\
  & \not\sim P(X)P(Y|X)
\end{align*}
where subscript $S$ stands for synthetic. Consequently, the expected \footnote{expected since $\hat{P}$ depends on the realization of synthetic sample $D_S$} KL-divergence between true target conditional distribution $P(Y|X)$ and the the learned distribution $\hat{P}(Y|X)$ can be decomposed as \citep{10.1162/089976698300017232}:
\begin{align*}
Error(\hat{P}) &= E_{D_S}\Bigl[KL\bigl(P\,\|\,\hat{P}\bigr)\Bigr] \\
&\approx KL(P\|P_S)
+ E_{D_S}\Bigl[KL\bigl(P_S\,\|\,\hat{P}\bigr)\Bigr]
\end{align*}
where the first term represents the irreducible approximation error coming from $P_S$, and the second terms is the estimation error coming from training. \paragraph{Interpretation} Crucially, the first term irreducible approximation error implies that no amount of synthetic labels can remove the systematic biases LLM annotators introduces, leading to performance plateau. The decomposition also helps explain the amplification of instability when training on synthetic labels. In addition to the usual finite sample estimation errors, in regions where $P_S(Y|X)$ is particularly off from $P(Y|X)$, even small fluctuations in the synthetic data can lead to larger estimation errors. Essentially, the estimation error can be amplified by the underlying approximation error, leading to more variance in performance across different training runs.
\begin{figure*}[t]
    \centering
    \includegraphics[width=0.9\linewidth]{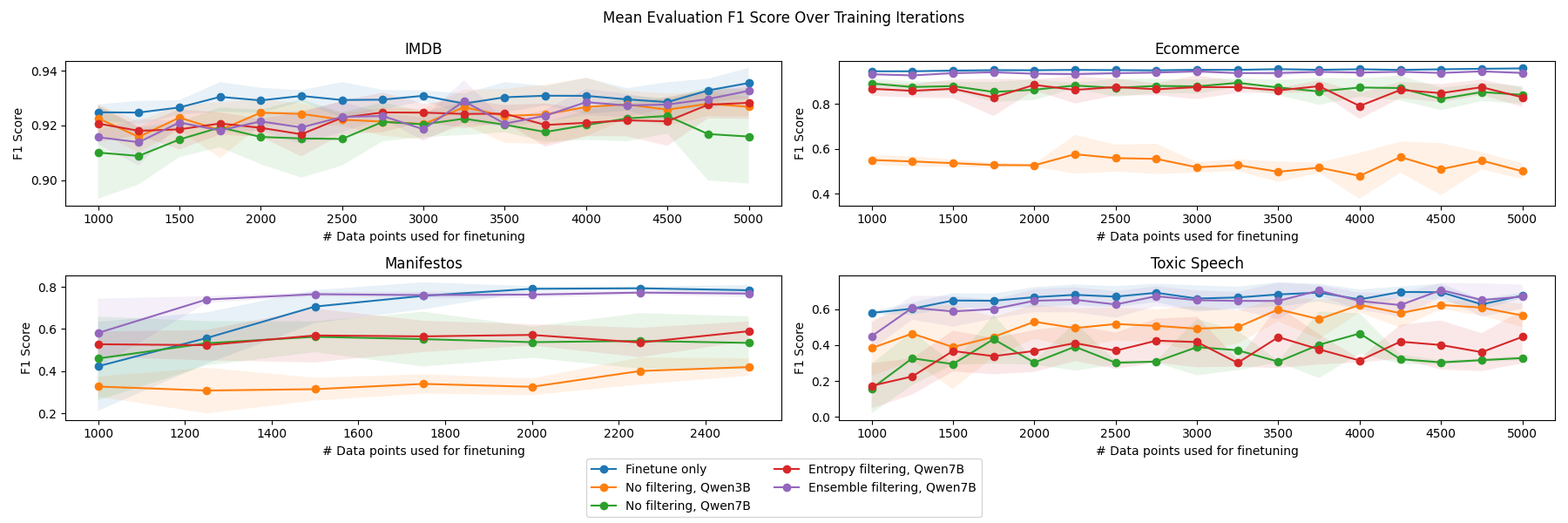}
    \caption{Performance as data point increases}
    \label{fig:enter-label}
\end{figure*}

\section{Mitigation Experiments}
\label{sec:mitigation setup}
As the theoretical framework suggests, the key driver of performance degradation is the divergence between the true conditional distribution $P(Y|X)$ and the LLM-generated distribution $P_S(Y|X)$. Intuitively, one way to mitigate this error is to filter out unreliable LLM-generated labels and increase the signal to noise ratio to control the error size. For any given input text $x$, we can try to control the size of error by mixing in true labels to obtain a better conditional distribution 
\begin{align*}
    P_F(y|X=x) = F(x) P_S(y|X=x) \\
    + (1-F(x)) P(y|X=x)
\end{align*}
where $F(x)$ is the data-dependent filtering function. In principle, one could parameterize F and treat it as a learnable function. However, in our low resource setup, we resort to computationally cheap heuristics.

We evaluate mitigation strategies using the 7B LLM annotator with RoBERTa-base, selected for its balance of performance and practical relevance.  
\paragraph{Entropy-Ranking Filtering}
For each input $x$, we compute the conditional entropy of the LLM's class predictions:
\begin{align*}
H(Y|X=x) = -\sum_{y \in \mathcal{Y}} p(y|x) \log p(y|x)
\end{align*}
where $p(y|x)$ is the LLM's predicted probability for class $y$. Entropy is commonly used as a baseline for assessing uncertainty \citep{huang2024surveyuncertaintyestimationllms}. We rank predictions and replace LLM annotations with gold labels if they are in the top $\alpha \in \{5, 25\}$ percent. Importantly, we are not using a fixed threshold to sidestep temperature scaling, and to account for the fact that many out-of-shelf LLMs are poorly calibrated \citep{desai-durrett-2020-calibration,zhu-etal-2023-calibration}.
Note that in binary classification, entropy ranking is equivalent to logits-ranking, which is the approach \citet{wang-etal-2021-want-reduce} took.

\paragraph{Consistency Ensemble}
Another simple fix is prompt LLM to generate multiple predictions with different demonstrations. The idea is that a robust prediction should not depend too much on the specific examples we provide in the prompt. We replace cases where predictions flip with human annotations.

\section{Mitigation Results}
\label{sec:mitigation results}
\begin{table*}[!ht]
\centering
\footnotesize
\setlength{\tabcolsep}{4pt}
\begin{tabular}{l l | ccc | ccc | ccc | ccc}
\hline
\multirow{3}{*}{Model} & \multirow{3}{*}{\begin{tabular}[c]{@{}l@{}}Label\\Type\end{tabular}} & \multicolumn{3}{c|}{IMDB} & \multicolumn{3}{c|}{Ecommerce} & \multicolumn{3}{c|}{Manifestos} & \multicolumn{3}{c}{Toxic} \\
& & $\mu_{acc}$ & $\mu_{f1}$ & $\alpha_{K}$ & $\mu_{acc}$ & $\mu_{f1}$ & $\alpha_{K}$ & $\mu_{acc}$ & $\mu_{f1}$ & $\alpha_{K}$ & $\mu_{acc}$ & $\mu_{f1}$ & $\alpha_{K}$ \\
& & $\sigma_{acc}$ & $\sigma_{f1}$ & $p_{uc}$ & $\sigma_{acc}$ & $\sigma_{f1}$ & $p_{uc}$ & $\sigma_{acc}$ & $\sigma_{f1}$ & $p_{uc}$ & $\sigma_{acc}$ & $\sigma_{f1}$ & $p_{uc}$ \\
\hline 
\multirow{6}{*}{\rotatebox{90}{RoBERTa-Base}}
& \multirow{2}{*}{Gold} & 93.84 & 93.82 & 90.6 & 96.26 & 96.23 & 96.69 & 83.56 & 79.38 & 84.30 & 91.13 & 75.24 & 84.08 \\
& & 0.28 & 0.29 & \cellcolor{red!60}90 & 0.25 & 0.22 & \cellcolor{red!60}95.24 & 0.69 & 0.54 & \cellcolor{red!60}82.04 & 0.23 & 1.01 & \cellcolor{red!60}88.50 \\
\cline{2-14}
& \multirow{2}{*}{Entropy} & 93.28 & 93.27 & 89.33 & 91.79 & 91.85 & 83.09 & 71.06 & 62.48 & 59.80 & 81.56 & 61.42 & 68.44 \\
& & 0.32 & 0.32 & \cellcolor{red!60}89 & 0.71 & 0.61 & \cellcolor{red!60}76.89 & 2.11 & 2.63 & \cellcolor{red!60}52.82 & 3.77 & 3.67 & \cellcolor{red!60}56.20 \\
\cline{2-14}
& \multirow{2}{*}{Ensemble} & 93.46 & 83.45 & 89.43 & 95.14 & 95.15 & 93.87 & 81.58 & 77.97 & 82.55 & 89.17 & 74.05 & 61.48 \\
& & 0.02 & 0.02 & \cellcolor{red!60}94.72 & 0.20 & 0.14 & \cellcolor{red!60}95.54 & 0.24 & 0.40 & \cellcolor{red!60}81.51 & 0.69 & 0.86 & \cellcolor{red!60}77.20 \\
\hline
\end{tabular}
\caption{Experimental results across different models, label types, and datasets. For each dataset, we report average accuracy $\mu_{acc}$, average macro F1-score $\mu_{f1}$, standard deviation of accuracy $\sigma_{acc}$, and standard deviation of macro F1 $\sigma_{f1}$. In addition, we compute Krippendorff's alpha $\alpha_{K}$ and the proportion of predictions that remain unchanged across experimental runs $p_{uc}$. All numbers are scaled up by 100 for ease of presentation.}
\label{tab:mitigation_results}
\end{table*}

\paragraph{Entropy-based Filtering} does not work well. While for IMDB, Ecommerce, Manifestos, entropy-based filtering stabilizes predictions to a certain. On Toxic, on the contrary, it leads to lowers the proportion of unchanged predictions $u_{pc}$ from $77.49$ to $56.20$. Given that entropy-based filtering is theoretical more appealing simple alternative simple uncertainty estimation heuristics including logits or log-probabilities, this does not bode well for prospect of having a cheap fix for the instability problem we identify. 

\paragraph{Consistency Ensemble} seems to work, but at a cost. Experiments suggests that consistency ensemble seems to manage to pick up many of LLM annotations that are distorting decision boundary for the classifiers. However, we need to be careful about this approach, however, because it requires multiple inferences on the same data point. For example, with 5 percent of the total unlabeled pool, 5-time ensemble means we are effectively performing inference on 25\% of the pool, which defeats the purpose of cost saving.

\section{Conclusion}
In this paper, we identify previously overlooked risks in using LLM-generated labels to train smaller text classifiers: performance drops, unstable predictions, and early plateaus in learning. These problems are worse for complex tasks and minority classes, which can amplify existing biases \cite{gallegos2024biasfairnesslargelanguage}. While we tested some fixes like filtering and ensembles, they only partially address these problems.
As \citet{chen2024surveylargelanguagemodels} warns about rushing to adopt LLMs without proper scrutiny, our results provide concrete evidence of risks in this specific use case. While using LLM-generated labels might work for simple tasks, we urge caution in critical applications.

\newpage
\section{Limitations and Ethical Considerations}
\paragraph{Limitations.}
One clear limitation is that our work does not offer a comprehensive solution to the problem we identified. While we explored a few heuristic mitigation strategies, we did not investigate more sophisticated approaches. For instance, our theoretical discussion suggests that using the embeddings as inputs to a simple ridge regression on a small validation set could help predict where the LLM is likely to make mistakes, thereby guiding targeted improvements through higher-quality annotations. However, given the scope of this short paper, we leave more in depth exploration of best strategies to  LLM-generated labels to text classification pipeline to future work. 

A second limitation stems from the rapid evolution of foundation models. As state-of-the-art models become increasingly capable of approximating the conditional distribution \(P(Y|X)\) arbitrarily well, our approach may become less relevant. Nonetheless, we welcome such advancements as they contribute positively to the field.

Finally, our theoretical analysis touches on the impact of approximation error, yet it lacks a rigorous exposition of how this error influences the variance and convergence rates of our estimates. Addressing this gap remains an important avenue for future research.

\paragraph{Ethical Considerations.}
We do not foresee significant ethical risks associated with our work. On the contrary, our paper cautions against the uncritical adoption of pipelines that utilize LLM-generated labels to fine-tune BERT-like models for classification. 

\paragraph{Use of AI} We acknowledge the use of artificial intelligence tools to assist with code debugging and prose refinement throughout this work.

\bibliography{llmbert_short, custom,anthology}

\begin{thebibliography}{37}
\providecommand{\natexlab}[1]{#1}
\providecommand{\url}[1]{\texttt{#1}}
\expandafter\ifx\csname urlstyle\endcsname\relax
  \providecommand{\doi}[1]{doi: #1}\else
  \providecommand{\doi}{doi: \begingroup \urlstyle{rm}\Url}\fi

\bibitem[Bauer et~al.(2024)Bauer, Trapp, Stenger, Leppich, Kounev, Leznik,
  Chard, and Foster]{bauer2024comprehensiveexplorationsyntheticdata}
André Bauer, Simon Trapp, Michael Stenger, Robert Leppich, Samuel Kounev, Mark
  Leznik, Kyle Chard, and Ian Foster.
\newblock Comprehensive exploration of synthetic data generation: A survey,
  2024.
\newblock URL \url{https://arxiv.org/abs/2401.02524}.

\bibitem[Chen et~al.(2022)Chen, Zheng, Awadallah, and
  Ji]{chen2022pathologiespretrainedlanguagemodels}
Hanjie Chen, Guoqing Zheng, Ahmed~Hassan Awadallah, and Yangfeng Ji.
\newblock Pathologies of pre-trained language models in few-shot fine-tuning,
  2022.
\newblock URL \url{https://arxiv.org/abs/2204.08039}.

\bibitem[Chen et~al.(2024)Chen, Ma, Zhang, Hao, Yan, Nourbakhsh, Yang, McAuley,
  Petzold, and Wang]{chen2024surveylargelanguagemodels}
Zhiyu~Zoey Chen, Jing Ma, Xinlu Zhang, Nan Hao, An~Yan, Armineh Nourbakhsh,
  Xianjun Yang, Julian McAuley, Linda Petzold, and William~Yang Wang.
\newblock A survey on large language models for critical societal domains:
  Finance, healthcare, and law, 2024.
\newblock URL \url{https://arxiv.org/abs/2405.01769}.

\bibitem[Colombo et~al.(2024)Colombo, Pires, Boudiaf, Culver, Melo, Corro,
  Martins, Esposito, Raposo, Morgado, and
  Desa]{colombo2024saullm7bpioneeringlargelanguage}
Pierre Colombo, Telmo~Pessoa Pires, Malik Boudiaf, Dominic Culver, Rui Melo,
  Caio Corro, Andre F.~T. Martins, Fabrizio Esposito, Vera~Lúcia Raposo, Sofia
  Morgado, and Michael Desa.
\newblock Saullm-7b: A pioneering large language model for law, 2024.
\newblock URL \url{https://arxiv.org/abs/2403.03883}.

\bibitem[Csanády et~al.(2024)Csanády, Muzsai, Vedres, Nádasdy, and
  Lukács]{csanády2024llambertlargescalelowcostdata}
Bálint Csanády, Lajos Muzsai, Péter Vedres, Zoltán Nádasdy, and András
  Lukács.
\newblock Llambert: Large-scale low-cost data annotation in nlp, 2024.
\newblock URL \url{https://arxiv.org/abs/2403.15938}.

\bibitem[Davidson et~al.(2017)Davidson, Warmsley, Macy, and
  Weber]{davidson2017automatedhatespeechdetection}
Thomas Davidson, Dana Warmsley, Michael Macy, and Ingmar Weber.
\newblock Automated hate speech detection and the problem of offensive
  language, 2017.
\newblock URL \url{https://arxiv.org/abs/1703.04009}.

\bibitem[Dell(2024)]{dell2024deeplearningeconomists}
Melissa Dell.
\newblock Deep learning for economists, 2024.
\newblock URL \url{https://arxiv.org/abs/2407.15339}.

\bibitem[Desai \& Durrett(2020)Desai and
  Durrett]{desai-durrett-2020-calibration}
Shrey Desai and Greg Durrett.
\newblock Calibration of pre-trained transformers.
\newblock In Bonnie Webber, Trevor Cohn, Yulan He, and Yang Liu (eds.),
  \emph{Proceedings of the 2020 Conference on Empirical Methods in Natural
  Language Processing (EMNLP)}, pp.\  295--302, Online, November 2020.
  Association for Computational Linguistics.
\newblock \doi{10.18653/v1/2020.emnlp-main.21}.
\newblock URL \url{https://aclanthology.org/2020.emnlp-main.21/}.

\bibitem[Devlin et~al.(2019)Devlin, Chang, Lee, and
  Toutanova]{devlin-etal-2019-bert}
Jacob Devlin, Ming-Wei Chang, Kenton Lee, and Kristina Toutanova.
\newblock {BERT}: Pre-training of deep bidirectional transformers for language
  understanding.
\newblock In Jill Burstein, Christy Doran, and Thamar Solorio (eds.),
  \emph{Proceedings of the 2019 Conference of the North {A}merican Chapter of
  the Association for Computational Linguistics: Human Language Technologies,
  Volume 1 (Long and Short Papers)}, pp.\  4171--4186, Minneapolis, Minnesota,
  June 2019. Association for Computational Linguistics.
\newblock \doi{10.18653/v1/N19-1423}.
\newblock URL \url{https://aclanthology.org/N19-1423/}.

\bibitem[Freitas(2024)]{freitas-2024-text}
Lucas Jos{\'e}~Gon{\c{c}}alves Freitas.
\newblock Text clustering applied to unbalanced data in legal contexts.
\newblock In Pablo Gamallo, Daniela Claro, Ant{\'o}nio Teixeira, Livy Real,
  Marcos Garcia, Hugo~Gon{\c{c}}alo Oliveira, and Raquel Amaro (eds.),
  \emph{Proceedings of the 16th International Conference on Computational
  Processing of Portuguese - Vol. 1}, pp.\  639--642, Santiago de Compostela,
  Galicia/Spain, March 2024. Association for Computational Lingustics.
\newblock URL \url{https://aclanthology.org/2024.propor-1.74/}.

\bibitem[Gallegos et~al.(2024)Gallegos, Rossi, Barrow, Tanjim, Kim,
  Dernoncourt, Yu, Zhang, and Ahmed]{gallegos2024biasfairnesslargelanguage}
Isabel~O. Gallegos, Ryan~A. Rossi, Joe Barrow, Md~Mehrab Tanjim, Sungchul Kim,
  Franck Dernoncourt, Tong Yu, Ruiyi Zhang, and Nesreen~K. Ahmed.
\newblock Bias and fairness in large language models: A survey, 2024.
\newblock URL \url{https://arxiv.org/abs/2309.00770}.

\bibitem[Gautam(2019)]{gautam_2019}
A.~Gautam.
\newblock E commerce text dataset (version - 2), 2019.
\newblock URL \url{https://doi.org/10.5281/zenodo.3355823}.

\bibitem[Golde et~al.(2023)Golde, Haller, Hamborg, Risch, and
  Akbik]{golde2023fabricator}
Jonas Golde, Patrick Haller, Felix Hamborg, Julian Risch, and Alan Akbik.
\newblock Fabricator: An open source toolkit for generating labeled training
  data with teacher llms.
\newblock In \emph{Proceedings of the 2023 Conference on Empirical Methods in
  Natural Language Processing: System Demonstrations}, pp.\  1--11, Singapore,
  2023. Association for Computational Linguistics.

\bibitem[Halterman \& Keith(2025)Halterman and
  Keith]{halterman2025codebookllmsevaluatingllms}
Andrew Halterman and Katherine~A. Keith.
\newblock Codebook llms: Evaluating llms as measurement tools for political
  science concepts, 2025.
\newblock URL \url{https://arxiv.org/abs/2407.10747}.

\bibitem[Heskes(1998)]{10.1162/089976698300017232}
Tom Heskes.
\newblock Bias/variance decompositions for likelihood-based estimators.
\newblock \emph{Neural Computation}, 10\penalty0 (6):\penalty0 1425--1433, 08
  1998.
\newblock ISSN 0899-7667.
\newblock \doi{10.1162/089976698300017232}.
\newblock URL \url{https://doi.org/10.1162/089976698300017232}.

\bibitem[Hopkins et~al.()Hopkins, Lelkes, and
  Wolken]{https://doi.org/10.1111/ajps.12875}
Daniel~J. Hopkins, Yphtach Lelkes, and Samuel Wolken.
\newblock The rise of and demand for identity-oriented media coverage.
\newblock \emph{American Journal of Political Science}, n/a\penalty0 (n/a).
\newblock \doi{https://doi.org/10.1111/ajps.12875}.
\newblock URL \url{https://onlinelibrary.wiley.com/doi/abs/10.1111/ajps.12875}.

\bibitem[Huang et~al.(2024)Huang, Yang, Zhang, Lee, and
  Wu]{huang2024surveyuncertaintyestimationllms}
Hsiu-Yuan Huang, Yutong Yang, Zhaoxi Zhang, Sanwoo Lee, and Yunfang Wu.
\newblock A survey of uncertainty estimation in llms: Theory meets practice,
  2024.
\newblock URL \url{https://arxiv.org/abs/2410.15326}.

\bibitem[Kumichev et~al.(2024)Kumichev, Blinov, Kuzkina, Goncharov, Zubkova,
  Zenovkin, Goncharov, and Savchenko]{Kumichev_2024}
Gleb Kumichev, Pavel Blinov, Yulia Kuzkina, Vasily Goncharov, Galina Zubkova,
  Nikolai Zenovkin, Aleksei Goncharov, and Andrey Savchenko.
\newblock \emph{MedSyn: LLM-Based Synthetic Medical Text Generation Framework},
  pp.\  215–230.
\newblock Springer Nature Switzerland, 2024.
\newblock ISBN 9783031703812.
\newblock \doi{10.1007/978-3-031-70381-2_14}.
\newblock URL \url{http://dx.doi.org/10.1007/978-3-031-70381-2_14}.

\bibitem[Kuo et~al.(2024)Kuo, Liao, Chao, Ma, and
  Cheng]{kuo2024llmgenerateddataequalrethinking}
Hsun-Yu Kuo, Yin-Hsiang Liao, Yu-Chieh Chao, Wei-Yun Ma, and Pu-Jen Cheng.
\newblock Not all llm-generated data are equal: Rethinking data weighting in
  text classification, 2024.
\newblock URL \url{https://arxiv.org/abs/2410.21526}.

\bibitem[Kwon et~al.(2023)Kwon, Li, Zhuang, Sheng, Zheng, Yu, Gonzalez, Zhang,
  and Stoica]{kwon2023efficient}
Woosuk Kwon, Zhuohan Li, Siyuan Zhuang, Ying Sheng, Lianmin Zheng, Cody~Hao Yu,
  Joseph~E. Gonzalez, Hao Zhang, and Ion Stoica.
\newblock Efficient memory management for large language model serving with
  pagedattention.
\newblock In \emph{Proceedings of the ACM SIGOPS 29th Symposium on Operating
  Systems Principles}, 2023.

\bibitem[Li et~al.(2023)Li, Zhu, Lu, and Yin]{li-etal-2023-synthetic}
Zhuoyan Li, Hangxiao Zhu, Zhuoran Lu, and Ming Yin.
\newblock Synthetic data generation with large language models for text
  classification: Potential and limitations.
\newblock In Houda Bouamor, Juan Pino, and Kalika Bali (eds.),
  \emph{Proceedings of the 2023 Conference on Empirical Methods in Natural
  Language Processing}, pp.\  10443--10461, Singapore, December 2023.
  Association for Computational Linguistics.
\newblock \doi{10.18653/v1/2023.emnlp-main.647}.
\newblock URL \url{https://aclanthology.org/2023.emnlp-main.647/}.

\bibitem[Liu et~al.(2024)Liu, Wei, Liu, Si, Zhang, Rao, Zheng, Peng, Yang,
  Zhou, and Dai]{liu2024best}
Ruibo Liu, Jerry Wei, Fangyu Liu, Chenglei Si, Yanzhe Zhang, Jinmeng Rao,
  Steven Zheng, Daiyi Peng, Diyi Yang, Denny Zhou, and Andrew~M. Dai.
\newblock Best practices and lessons learned on synthetic data.
\newblock In \emph{First Conference on Language Modeling}, 2024.
\newblock URL \url{https://openreview.net/forum?id=OJaWBhh61C}.

\bibitem[Liu et~al.(2019)Liu, Ott, Goyal, Du, Joshi, Chen, Levy, Lewis,
  Zettlemoyer, and Stoyanov]{liu2019robertarobustlyoptimizedbert}
Yinhan Liu, Myle Ott, Naman Goyal, Jingfei Du, Mandar Joshi, Danqi Chen, Omer
  Levy, Mike Lewis, Luke Zettlemoyer, and Veselin Stoyanov.
\newblock Roberta: A robustly optimized bert pretraining approach, 2019.
\newblock URL \url{https://arxiv.org/abs/1907.11692}.

\bibitem[Maas et~al.(2011)Maas, Daly, Pham, Huang, Ng, and
  Potts]{maas-EtAl:2011:ACL-HLT2011}
Andrew~L. Maas, Raymond~E. Daly, Peter~T. Pham, Dan Huang, Andrew~Y. Ng, and
  Christopher Potts.
\newblock Learning word vectors for sentiment analysis.
\newblock In \emph{Proceedings of the 49th Annual Meeting of the Association
  for Computational Linguistics: Human Language Technologies}, pp.\  142--150,
  Portland, Oregon, USA, June 2011. Association for Computational Linguistics.
\newblock URL \url{http://www.aclweb.org/anthology/P11-1015}.

\bibitem[Mohamed~Serouis \& Sèdes(2024)Mohamed~Serouis and Sèdes]{10825193}
Ibrahim Mohamed~Serouis and Florence Sèdes.
\newblock Leveraging llms for fair data labeling and validation in
  crowdsourcing environments [vision paper].
\newblock In \emph{2024 IEEE International Conference on Big Data (BigData)},
  pp.\  468--472, 2024.
\newblock \doi{10.1109/BigData62323.2024.10825193}.

\bibitem[Müller(2020)]{DVN/7NP2XH_2020}
Stefan Müller.
\newblock {Replication Data for: The Temporal Focus of Campaign Communication},
  2020.
\newblock URL \url{https://doi.org/10.7910/DVN/7NP2XH}.

\bibitem[Pangakis \& Wolken(2024{\natexlab{a}})Pangakis and
  Wolken]{pangakis-wolken-2024-knowledge}
Nicholas Pangakis and Sam Wolken.
\newblock Knowledge distillation in automated annotation: Supervised text
  classification with {LLM}-generated training labels.
\newblock In Dallas Card, Anjalie Field, Dirk Hovy, and Katherine Keith (eds.),
  \emph{Proceedings of the Sixth Workshop on Natural Language Processing and
  Computational Social Science (NLP+CSS 2024)}, pp.\  113--131, Mexico City,
  Mexico, June 2024{\natexlab{a}}. Association for Computational Linguistics.
\newblock \doi{10.18653/v1/2024.nlpcss-1.9}.
\newblock URL \url{https://aclanthology.org/2024.nlpcss-1.9/}.

\bibitem[Pangakis \& Wolken(2024{\natexlab{b}})Pangakis and
  Wolken]{pangakis2024keepinghumansloophumancentered}
Nicholas Pangakis and Samuel Wolken.
\newblock Keeping humans in the loop: Human-centered automated annotation with
  generative ai, 2024{\natexlab{b}}.
\newblock URL \url{https://arxiv.org/abs/2409.09467}.

\bibitem[Shumailov et~al.(2024{\natexlab{a}})Shumailov, Shumaylov, Zhao, Gal,
  Papernot, and Anderson]{shumailov2024curserecursiontraininggenerated}
Ilia Shumailov, Zakhar Shumaylov, Yiren Zhao, Yarin Gal, Nicolas Papernot, and
  Ross Anderson.
\newblock The curse of recursion: Training on generated data makes models
  forget, 2024{\natexlab{a}}.
\newblock URL \url{https://arxiv.org/abs/2305.17493}.

\bibitem[Shumailov et~al.(2024{\natexlab{b}})Shumailov, Shumaylov, Zhao,
  et~al.]{shumailov2024collapse}
Ilia Shumailov, Zakhar Shumaylov, Yiren Zhao, et~al.
\newblock Ai models collapse when trained on recursively generated data.
\newblock \emph{Nature}, 631:\penalty0 755--759, 2024{\natexlab{b}}.
\newblock \doi{10.1038/s41586-024-07566-y}.

\bibitem[Smolyak et~al.(2024)Smolyak, Bjarnadóttir, Crowley, and
  Agarwal]{smolyak2024large}
Daniel Smolyak, Margrét~V Bjarnadóttir, Kathy Crowley, and Ritu Agarwal.
\newblock Large language models and synthetic health data: progress and
  prospects.
\newblock \emph{JAMIA Open}, 7\penalty0 (4):\penalty0 ooae114, 12 2024.
\newblock ISSN 2574-2531.
\newblock \doi{10.1093/jamiaopen/ooae114}.
\newblock URL \url{https://doi.org/10.1093/jamiaopen/ooae114}.

\bibitem[Sun et~al.(2020)Sun, Qiu, Xu, and
  Huang]{sun2020finetuneberttextclassification}
Chi Sun, Xipeng Qiu, Yige Xu, and Xuanjing Huang.
\newblock How to fine-tune bert for text classification?, 2020.
\newblock URL \url{https://arxiv.org/abs/1905.05583}.

\bibitem[Wang et~al.(2021)Wang, Liu, Xu, Zhu, and
  Zeng]{wang-etal-2021-want-reduce}
Shuohang Wang, Yang Liu, Yichong Xu, Chenguang Zhu, and Michael Zeng.
\newblock Want to reduce labeling cost? {GPT}-3 can help.
\newblock In Marie-Francine Moens, Xuanjing Huang, Lucia Specia, and Scott
  Wen-tau Yih (eds.), \emph{Findings of the Association for Computational
  Linguistics: EMNLP 2021}, pp.\  4195--4205, Punta Cana, Dominican Republic,
  November 2021. Association for Computational Linguistics.
\newblock \doi{10.18653/v1/2021.findings-emnlp.354}.
\newblock URL \url{https://aclanthology.org/2021.findings-emnlp.354/}.

\bibitem[Willard \& Louf(2023)Willard and Louf]{willard2023efficient}
Brandon~T Willard and R{\'e}mi Louf.
\newblock Efficient guided generation for llms.
\newblock \emph{arXiv preprint arXiv:2307.09702}, 2023.

\bibitem[Wolf et~al.(2020)Wolf, Debut, Sanh, Chaumond, Delangue, Moi, Cistac,
  Rault, Louf, Funtowicz, Davison, Shleifer, von Platen, Ma, Jernite, Plu, Xu,
  Le~Scao, Gugger, Drame, Lhoest, and Rush]{wolf-etal-2020-transformers}
Thomas Wolf, Lysandre Debut, Victor Sanh, Julien Chaumond, Clement Delangue,
  Anthony Moi, Pierric Cistac, Tim Rault, Remi Louf, Morgan Funtowicz, Joe
  Davison, Sam Shleifer, Patrick von Platen, Clara Ma, Yacine Jernite, Julien
  Plu, Canwen Xu, Teven Le~Scao, Sylvain Gugger, Mariama Drame, Quentin Lhoest,
  and Alexander Rush.
\newblock Transformers: State-of-the-art natural language processing.
\newblock In Qun Liu and David Schlangen (eds.), \emph{Proceedings of the 2020
  Conference on Empirical Methods in Natural Language Processing: System
  Demonstrations}, pp.\  38--45, Online, October 2020. Association for
  Computational Linguistics.
\newblock \doi{10.18653/v1/2020.emnlp-demos.6}.
\newblock URL \url{https://aclanthology.org/2020.emnlp-demos.6/}.

\bibitem[Yang et~al.(2024)Yang, Yang, Zhang, Hui, Zheng, Yu, Li, Liu, Huang,
  Wei, Lin, Yang, Tu, Zhang, Yang, Yang, Zhou, Lin, Dang, Lu, Bao, Yang, Yu,
  Li, Xue, Zhang, Zhu, Men, Lin, Li, Xia, Ren, Ren, Fan, Su, Zhang, Wan, Liu,
  Cui, Zhang, and Qiu]{qwen2.5}
An~Yang, Baosong Yang, Beichen Zhang, Binyuan Hui, Bo~Zheng, Bowen Yu,
  Chengyuan Li, Dayiheng Liu, Fei Huang, Haoran Wei, Huan Lin, Jian Yang,
  Jianhong Tu, Jianwei Zhang, Jianxin Yang, Jiaxi Yang, Jingren Zhou, Junyang
  Lin, Kai Dang, Keming Lu, Keqin Bao, Kexin Yang, Le~Yu, Mei Li, Mingfeng Xue,
  Pei Zhang, Qin Zhu, Rui Men, Runji Lin, Tianhao Li, Tingyu Xia, Xingzhang
  Ren, Xuancheng Ren, Yang Fan, Yang Su, Yichang Zhang, Yu~Wan, Yuqiong Liu,
  Zeyu Cui, Zhenru Zhang, and Zihan Qiu.
\newblock Qwen2.5 technical report.
\newblock \emph{arXiv preprint arXiv:2412.15115}, 2024.

\bibitem[Zhu et~al.(2023)Zhu, Xu, Wang, Zhang, and
  Mao]{zhu-etal-2023-calibration}
Chiwei Zhu, Benfeng Xu, Quan Wang, Yongdong Zhang, and Zhendong Mao.
\newblock On the calibration of large language models and alignment.
\newblock In Houda Bouamor, Juan Pino, and Kalika Bali (eds.), \emph{Findings
  of the Association for Computational Linguistics: EMNLP 2023}, pp.\
  9778--9795, Singapore, December 2023. Association for Computational
  Linguistics.
\newblock \doi{10.18653/v1/2023.findings-emnlp.654}.
\newblock URL \url{https://aclanthology.org/2023.findings-emnlp.654/}.

\end{thebibliography}
\bibliographystyle{colm2025_conference}

\appendix
\section{Dataset Descriptions}
\label{sec:datasets}

\paragraph{IMDB} The Stanford Large Movie Review Dataset \citep{maas-EtAl:2011:ACL-HLT2011}, IMDB for short, needs no introduction among NLP practitioners. 
\paragraph{E-commerce} \citep{gautam_2019} The Ecommerce dataset contains 50,425 product listings scraped from Indian ecommerce platforms, consisting of product titles and descriptions. Each item is categorized into one of four classes: Electronics, Household, Books, or Clothing and Accessories. The dataset is slightly imbalanced across these four classes, with each product represented by its textual description. 
\paragraph{Manifestos} \citep{DVN/7NP2XH_2020} The Manifesto Project dataset comprises annotated political texts, including party election manifestos from 50+ countries, labeled with policy positions and topics. We focus on the English-language subset, which includes over 4,000 documents annotated at the sentence level. Each sentence is categorized into one of 56 policy areas (e.g., "Environment," "Education"). The dataset is widely used for political text analysis and multi-label classification tasks. We preprocess the text to remove metadata and retain only sentences with unambiguous policy labels.
\paragraph{Toxic speech} \citep{davidson2017automatedhatespeechdetection} This dataset contains 24,802 tweets annotated via crowd-sourcing into three categories: \textit{hate speech}, \textit{offensive language}, or \textit{neither}. Tweets were collected using a crowd-sourced lexicon of hate speech keywords, and annotations emphasize distinguishing hate speech (targeted attacks on protected groups) from general offensiveness. The dataset is imbalanced, with most tweets labeled as offensive. Racist and homophobic content is more reliably classified as hate speech, while sexist remarks are often misclassified as merely offensive. We use this dataset to evaluate nuanced hate speech detection, focusing on precision-recall trade-offs.
To reduce environmental impacts, we limit the number of data points for train to up to 5000 for all datasets and shrink the size of test datasets with $<=2000$ by randomly drawing from existing test sets. 

\section{Fine-tuning Details}
\label{sec:slm details}
We employ Huggingface's pre-trained weights for both BERT \citep{devlin-etal-2019-bert} and RoBERTa \citep{liu2019robertarobustlyoptimizedbert} as provided in the Transformers library \citep{wolf-etal-2020-transformers}. We conduct full fine-tuning of the pre-trained language models following \citet{sun2020finetuneberttextclassification}, without freezing any pre-trained layers. The classification head consists of a dropout layer (set at the default value 0.1) followed by a linear layer that maps the [CLS] token representation to dimension of the target label space. 

While extensive hyperparameter tuning could potentially yield better performance, we prioritize consistent experimental conditions across datasets to isolate the effects of synthetic labels on performance stability. As a result, our baseline performance on gold-label fine-tuning may be slightly below state-of-the-art, but provides a fair foundation for comparative analysis.

Training runs for 3 epochs with a batch size of 16 for training and 32 for evaluation. We use the AdamW optimizer with a learning rate of 2e-5 and weight decay of 0.01. A linear learning rate scheduler with a warmup ratio of 0.05 is applied. The best checkpoint is selected based on validation F1 score, with a maximum of 2 checkpoints saved during training to conserve storage. All experiments use mixed-precision training (FP16) and are conducted on a single NVIDIA RTX 8000 GPU. \\

\section{LLM Annotation Details}
\label{sec:llm details}
We utilize \texttt{vLLM} \citep{kwon2023efficient} for improved memory efficiency and to better simulate a production environment. In addition, guided decoding \citep{willard2023efficient} is imposed to ensure that the outputs follow a consistent format. In particular, the model is constrained to generate only two tokens: the first token is the predicted class token (with labels mapped to integers) and the second token is the end-of-sequence (\texttt{<EOS>}) marker. The annotation pipeline uses a structured prompt template that puts together a task description, label description, demonstrations (randomly drawn from training datasets), and input text as follows:
\begin{verbatim}
### Instruction ###
{task description}
Respond with only the label name, nothing else.
### Available Labels ###
{label description}
### Examples ###
{demonstrations}
### Input ###
Text to classify: {input_text}
### Output ###
Label:
\end{verbatim}
\newpage
\begin{table}[t]
    \centering
    \small
    \begin{tabular}{|l|p{8cm}|p{5cm}|}
    \hline
    \textbf{Dataset} & \textbf{Task Description} & \textbf{Label Mapping} \\
    \hline
    IMDB & 
    You are an AI assistant specializing in sentiment analysis of movie reviews. You are going to help classify movie reviews as positive or negative. & 
    \{"0": ``negative'', "1": ``positive''\} \\
    \hline
    Ecommerce & 
    You are an AI assistant and you are very good at doing ecommerce products classification. You are going to help a customer to classify the products on the ecommerce website. & 
    \{"0": ``books'', "1": ``clothing \& accessories'', "2": ``electronics'', "3": ``household''\} \\
    \hline
    Manifestos & 
    You are an AI assistant specializing in classifying the temporal alignment of political party manifestos.You are going to help classify political party manifestos as about the future, the present, or the past. & 
    \{"0": ``present'', "1": ``future'', "2": ``past''\} \\
    \hline
    Toxic & 
    You are an AI assistant specializing in detecting hate speech and offensive language. You are going to help classify tweets as hate speech, offensive language, or neither. & 
    \{"0": ``hate speech'', "1": ``offensive language'', "2": ``neither''\} \\
    \hline
    \end{tabular}
    \caption{Task specifications for various datasets.}
    \label{tab:task_specs}
\end{table}

\end{document}